\documentclass[sigconf]{acmart}

\AtBeginDocument{%
  \providecommand\BibTeX{{%
    \normalfont B\kern-0.5em{\scshape i\kern-0.25em b}\kern-0.8em\TeX}}}
    
\acmConference[SSL '21]{SSL '121: The International Workshop on Self-Supervised Learning for the Web}{April 19--23, 2021}{SSL, LJ}
\acmBooktitle{SSL '21: The International Workshop on Self-Supervised Learning for the Web,
  April 19--23, 2021, SSL, LJ}
\acmPrice{15.00}
\acmISBN{978-1-4503-XXXX-X/18/06}

\usepackage{bm}
\usepackage{multirow}
\usepackage{color}

\newcommand{\model}{\textsc{SelfGNN}}
\newcommand{\cluster}{\textsc{ClusterSelfGNN}}
\newcommand{\bn}{\textsc{BatchNorm}}
\newcommand{\lrn}{\textsc{LayerNorm}}
\newcommand{\pagerank}{\textsc{PageRank}}

\newcommand{\ppr}{\textsc{PPR}}
\newcommand{\hk}{\textsc{HK}}
\newcommand{\katz}{\textsc{Katz}}
\newcommand{\splt}{\textsc{Split}}
\newcommand{\std}{\textsc{Standard}}
\newcommand{\paste}{\textsc{Paste}}
\newcommand{\ldp}{\textsc{LDP}}

\newcommand{\selfppr}{\textsc{SelfPPR}}
\newcommand{\selfhk}{\textsc{SelfHK}}
\newcommand{\selfkatz}{\textsc{SelfKatz}}

\newcommand{\selfbfa}{\textsc{SelfBFA}}

\newcommand{\gcn}{\textsc{GCN}}
\newcommand{\gat}{\textsc{GAT}}
\newcommand{\sage}{\textsc{GraphSage}}
\newcommand{\clustergcn}{\textsc{ClusterGCN}}
\newcommand{\saint}{\textsc{GraphSaint}}
\newcommand{\mvsgnn}{\textsc{MVS-GNN}}
\newcommand{\metis}{\textsc{METIS}}

\newcommand{\mvgrl}{\textsc{MvGrl}}

\newcommand{\mvgrlppr}{\textsc{MvGrlPPR}}
\newcommand{\mvgrlhk}{\textsc{MvGrlHK}}
\newcommand{\mvgrlkatz}{\textsc{MvGrlKatz}}
\newcommand{\dgi}{\textsc{DGI}}

\def\vs{{\bm{s}}}

\def\vx{{\bm{x}}}

\def\mA{{\bm{A}}}

\def\mD{{\bm{D}}}

\def\mH{{\bm{H}}}
\def\mI{{\bm{I}}}

\def\mW{{\bm{W}}}
\def\mX{{\bm{X}}}

\def\gL{{\mathcal{L}}}

\def\sR{{\mathbb{R}}}

\begin{document}

\title{\model: Self-supervised Graph Neural Networks without explicit negative sampling}

\author{Zekarias T. Kefato}
\affiliation{%
  \institution{KTH Royal Institute of Technology}
  \city{Stockholm}
  \country{Sweden}}
\email{zekarias@kth.se}

\author{Sarunas Girdzijauskas}
\affiliation{%
  \institution{KTH Royal Institute of Technology}
  \city{Stockholm}
  \country{Sweden}}
\email{sarunasg@kth.se}

\renewcommand{\shortauthors}{Kefato and Girdzijauskas}


\begin{abstract}
Real world data is mostly unlabeled or only few instances are labeled. 
Manually labeling data is a very expensive and daunting task.
This calls for unsupervised learning techniques that are powerful 
enough to achieve comparable results as semi-supervised/supervised techniques.
Contrastive self-supervised learning has emerged as a powerful direction, in some cases outperforming supervised techniques.

In this study, we propose, {\model}, a novel contrastive self-supervised graph neural network (GNN) without relying on explicit contrastive terms.
We leverage Batch Normalization, which introduces implicit contrastive terms, without sacrificing performance. 
Furthermore, as data augmentation is key in contrastive learning, we introduce four feature augmentation (FA) techniques for graphs.
Though graph topological augmentation (TA) is commonly used, our empirical findings show that FA perform as good as TA. Moreover, FA incurs no computational overhead, unlike TA, which often has $O(N^3)$ time complexity, $N$ -- number of nodes.

Our empirical evaluation on seven publicly available real-world data shows that, {\model} is powerful and leads to a performance comparable with SOTA supervised GNNs and always better than SOTA semi-supervised and unsupervised GNNs.
The source code is available at \href{https://github.com/zekarias-tilahun/SelfGNN}{https://github.com/zekarias-tilahun/SelfGNN}.

\end{abstract}



\maketitle


\section{Introduction}
\label{sec:intro}
Self-Supervised learning (SSL) has emerged as powerful representation learning paradigm bridging the gap between unsupervised and supervised learning methods.
Recent developments across different fields, such as Computer Vision (CV) and Natural Language Processing (NLP), achieve promising results using SSL techniques in some cases performing better than supervised methods~\cite{pmlr-v119-chen20j,grill2020bootstrap}.
These methods are powered by the so-called \emph{contrastive learning} (CL), which learns by contrasting augmented view of positive and negative objects.
That is, given augmented views of the same objects, the CL framework learns by maximizing the compatibility between the respective representation of the views and pushing them apart from representations of a contrastive term (negative sample).

Recent studies~\cite{grill2020bootstrap,chen2020exploring,richemond2020byol} in CV propose methods that do not require explicit contrastive terms, while achieving better performance than those with explicit contrastive terms.
As a result, now it is fairly understood that for some problems in CV, contrastive learning using either explicit or implicit negative samples performs at least as well as supervised methods~\cite{grill2020bootstrap,robinson2021contrastive,richemond2020byol}.
However, it is not yet clear whether contrastive terms should be explicit in GNNs.

Motivated by these findings, this study proposes {\model}, a contrastive self-supervised algorithm for graph neural networks with implicit contrastive terms.
{\model} imitates a Siamese network~\cite{10.5555/2987189.2987282}, which has been widely used in recent contrastive self-supervised learning methods~\cite{pmlr-v119-chen20j,grill2020bootstrap,chen2020exploring,richemond2020byol,srinivas2020curl,he2020momentum,robinson2021contrastive,hassani2020contrastive}.
While {\model} bares some resemblance to~\cite{grill2020bootstrap,chen2020exploring}, its unique characteristic is in contrast to virtually all SSL methods for graphs that require explicit negative sampling.
{\model} uses implicit negative samples that come as a result of batch normalization~\cite{fetterman2020understandingbyol,richemond2020byol}~\footnote{Update:~\cite{richemond2020byol} empirically shows that batch norm mainly solves improper initialization than preventing collapse or introducing implicit contrastive terms.}.
Furthermore, {\model} is agnostic to the type of GNN.

As data augmentation is the key component of CL, we introduce four node feature augmentation (FA) strategies, of which some are motivated by related techniques in CV.
Though, previously focus has been given to topological augmentation (TA), our empirical finding shows that both TA and FA provide competitive results.
However, TA techniques normally require expensive matrix operation with $O(N^3)$ time complexity, while, the proposed FA techniques are simply constant time operations.

The key contribution of this study is the introduction of a new class of self-supservised approaches for GNNs that requires no explicit negative sampling.
In addition, we have introduced four FA techniques that offer competitive performance as TA.

Extensive empirical evaluation using seven real-world datasets demonstrate that {\model} achieves comparable performance with supervised GNNs.
Our experiments show that compared to semi-supervised and unsupervised GNN methods it consistently achieves better performance.


\section{Related Work}
\label{sec:related_work}
As this study focuses on the intersection of Graph Neural Networks (GNN) and Self-Supervised Learning (SSL), in the following we cover related studies from both topics.

\subsection{Graph Neural Networks}
Graph Neural Networks represent a family of powerful graph representation learning algorithms~\cite{10.1145/2623330.2623732,10.1145/2736277.2741093,kipf2017semisupervised,10.5555/3294771.3294869,chen2018fastgcn,hassani2020contrastive,kefato2020gossip}.
Though they come in different flavors, most of them share an essential framework where representations are learned by propagating node features along the topology of the graph.
Unlike other neural networks, such as MLP and CNN, GNNs enable us to construct arbitrary architectures defined by the graph topology.
That is, the graph topology itself is what defines the neural network architecture.
Therefore, applying consecutive layers of GNNs allows node information to propagate to distant neighbors.
Due to the vast amount of literature on GNNs, we shall cover a selected few only.

A basic propagation rule has the form defined in Eq.~\ref{eq:gcn_propagation_rule}.

\begin{equation}\label{eq:gcn_propagation_rule}
    \mX^{l + 1} = \texttt{ReLU} (\mH\mX^l\mW^l)
\end{equation}
where, $\mH \in \sR^{N \times N}$ is a particular materialization of the topology, for example a symmetrically renormalized adjacency matrix, $\mA$ (${\mH = \texttt{symrnorm}(\mA)}$), $N$ number of nodes,  ${\mW^l \in \sR^{F^{l-1} \times F^l}}$ and $\mX^l \in \sR^{N \times F^l}$ are the parameters and activations of the $l^{th}$ layer, respectively, $F^l$ is the number of units of the $l^{th}$ layer. 
$\mW^l$ is shared by all nodes and this variant is commonly referred to as Graph Convolutional Network (GCN)~\cite{kipf2017semisupervised}.
A key limitation of GCN is that, it gives equal importance to features when aggregating. 
Graph Attention Network (GAT)~\cite{velickovic2018graph} is later proposed to address this by introducing learnable attention weights of neighborhood features according to their importance.

GCN and GAT require full-batch training, that is, the entire graph ($\mH$) should be loaded to memory, which makes them to be transductive and not scalable to large networks.
GraphSage~\cite{10.5555/3294771.3294869} alleviates these limitations and proposed a technique based on neighborhood sampling.
There, every node apriori specifies a selected few nodes that propagate information.
Nonetheless, this has introduced a memory overflow issue for a deeper model, which comes as a result of neighborhood explosion problem. 
Though, improved methods through layer sampling were proposed by other studies~\cite{chen2018fastgcn,zou2019layerdependent}, the problem has not been alleviated completely.

Followup studies propose subgraph sampling methods, using clustering (\clustergcn~\cite{10.1145/3292500.3330925}) and graph sampling (\saint~\cite{zeng2020graphsaint}, \mvsgnn~\cite{10.1145/3394486.3403192}) that are both scalable and do not suffer from the neighborhood explosion problem.
These studies pioneer a more efficient mini-batch training than their predecessors, using subgraphs as batch, and they are agnostic to the type of GNN.
The key idea is that, the subgraphs will be considered by their own merit and any type of GNN model can be applied on top of them.

\subsection{Self-Supervised Learning}
Self-supervised learning (SSL) has catalyzed representation learning across disciplines, such as CV, NLP, and GRL, owing to the critical problem of data labeling it tackles.
Real-world data is rarely labeled; and often, as a result of its sheer volume, labeling it is very expensive.
SSL has emerged as a standard representation learning approach, reaching to and sometimes surpassing the level of pretrained models learned through supervision~\cite{pmlr-v119-chen20j,grill2020bootstrap,chen2020exploring,robinson2021contrastive,hassani2020contrastive}.  

Powerful models, like BERT~\cite{devlin2019bert} and GPT~\cite{brown2020language}, has revolutionized SSL through the so-called \emph{pre-training and fine-tuning} framework. 
That is, they pretrain a model based on Transformers on a cheaper pretext task and will fine-tune it on a specifically difficult/expensive task of interest.
Difficulty or expensiveness here refers to the lack of sufficient labeled data or in general training data.

Alternatively, Contrastive Learning (CL) has gained popularity in the CV community as a variant of SSL for visual representation~\cite{pmlr-v119-chen20j,grill2020bootstrap,chen2020exploring,he2020momentum,srinivas2020curl}.
CL is based on data augmentation of a self and cotrastive term, where learning is carried out by maximizing similarities of the representations of the augmented views of the same object and minimizing similarity with respect to the conrastive object.

GRL has entertained both the Pre-train fine-tune~\cite{hu2020strategies,10.1145/3394486.3403237,10.1145/3394486.3403168} and CL~\cite{Velickovic2019DeepGI,hassani2020contrastive} paradigms. 
In the former, in general they closely follow the strategies of BERT and GPT families.
The later methods rely on augmented views of the original graph topology and design an objective that leverages the augmented view for contrasting.
\dgi~\cite{Velickovic2019DeepGI} is one CL method that uses a corrupted view of the original graph as an augmentation.
Another method, \mvgrl~\cite{hassani2020contrastive} on the other hand builds a higher-order network as an augmentation.
Though our approach can fall under CL methods, however unlike other methods we do not require explicit negative sampling.


\section{Method}
\label{sec:method}
We start by introducing the preliminaries upon which we build further discussion of the proposal.

\subsection{Preliminaries}
We consider an undirected graph, $G = (V, E, \mX)$, with a set of nodes $V$, and set of edges $E$, where $N = \vert V \vert $ and $M = \vert E \vert$.
${\mX \in \sR^{N \times F}}$ is a node feature matrix, where $F$ is the number of features.
The adjacency matrix representation is given by $\mA = [0, 1]^{N \times N}$, where $\mA[i, j] = 1$ if $(i, j) \in E$ and $0$ otherwise.
A diagonal matrix ${\mD \in \sR^{N \times N}}$ defines the degree distribution of $\mA$, and ${\mD[i, i] = \sum_{j=0}^{N} \mA[i,j]}$.
Finally, we define a symmetrically renormalized adjacency matrix as ${\Tilde{A} = \mD^{-1/2}(\mA + \mI)\mD^{-1/2}}$, and $\mI$ is the identity matrix.

In essence, CL brings together representations of two augmented views of the same object and repulses them apart from the representation of augmented views of a contrastive object, negative sample.


A question that has been carefully investigated in CV and yet to be understood clearly in graph representation learning is whether we can drop negative samples or use them implicitly without compromising performance~\cite{grill2020bootstrap}.
In CV, it has been shown that contrastive learning with implicit negative sampling~\cite{grill2020bootstrap} or hard negative sampling~\cite{robinson2021contrastive} can achieve as good a result as a pre-training with supervision.
This motivates us to investigate the former direction and shed some insights by avoiding explicit negative sampling in GRL, as the later is shown to be useful in GRL.
As a result, we rely on batch normalization (\bn) that introduces implicit negative samples~\cite{tian2020understanding}.

Note that, existing self-supervised methods for graphs rely on one or more of the following techniques:
\begin{enumerate}
    \item Pre-training and fine-tuning paradigm~\cite{hu2020strategies,10.1145/3394486.3403237}
    \item CL with explicit negative samples~\cite{hassani2020contrastive}
    \item Self-supervised learning with few class labels~\cite{sun2020multistage}
\end{enumerate}
In contrast with 1 and 2, our approach requires only positive augmented views as input, the contrastive term will be provided through a \bn, and in contrast with 3, no label information, whatsoever, is used during training.


As a standard contrastive learning framework requires a data augmentation strategy, we first introduce the graph data augmentation techniques we use in this study.

\subsection{Graph Data Augumentations}
Data augmentation in other domains, such as CV, has been extensively studied. 
As a result a set of standard augmentation techniques, such as cropping, rotating, blurring are commonly used.
However, there are no established data augmentation techniques for graphs~\cite{hassani2020contrastive}.
In this study, we propose two family of data augmentation techniques based on topology and features.
To the best of our knowledge, this is the first study proposing graph data augmentation techniques based on both topology and features..

\subsubsection*{Topology augmentation}
The aim of a topological data augmentation is to uncover a different topological view of the original graph by exploiting the properties of the graph structure.
A popular augmentation technique uses a higher-order network that is computed through a general graph diffusion process~\cite{klicpera2019diffusion}.
Employing high-order networks obtained through a diffusion process have been shown to improve GNNs performance~\cite{klicpera2019diffusion}.
In this study, we use two flavors of the popular {\pagerank} algorithm, which are {\pagerank} based on rooted random walks (commonly referred as personalized \pagerank--PPR), and heat-kernel (HK) \pagerank~\cite{Chung2007TheHK}, given in equations~\ref{eq:ppr_diffusion},~\ref{eq:hk_diffusion} respectively. 
We propose a third high-order network construction technique based on Katz-index as shown in Eq.~\ref{eq:katz_diffusion}. 
\begin{equation}\label{eq:ppr_diffusion}
    \mH^{PPR} = \alpha (\mI - (1 - \alpha) \Tilde{A})^{-1}
\end{equation}
\begin{equation}\label{eq:hk_diffusion}
    \mH^{HK} = \exp(t\mA\mD^{-1} - t)
\end{equation}
\begin{equation}\label{eq:katz_diffusion}
    \mH^{katz} = (I - \beta \Tilde{\mA})^{-1}\beta\Tilde{A}
\end{equation}
where $\alpha$ is a teleportation probability and $t$ is diffusion time, and $\beta$ is a decay parameter.
Katz-index is the weighted sum of the set of all paths between a pair of nodes, paths are penalized according to their length.
The attenuation factor ($\beta$) governs the penalization.

\subsubsection*{Feature Augmentation}
While graph data augmentation in SOTA largely focuses on the topological view, little attention has been given to feature augmentation.
Furthermore, a study~\cite{hassani2020contrastive} has argued against feature augmentation techniques, such as masking, and adding noise.
In this study, we propose the following feature augmentation techniques other than masking and noising. 
\begin{enumerate}
    \item \emph{Split}: the first feature augmentation technique is inspired by cropping for image augmentation, and it creates an augmented view of the features by splitting them into two as $\mX = \mX[:, :F/2]$ and $\mX' = \mX[:, F/2:]$
    That is, the first half of the feature dimension is used to build one view and while the remaining half is used to build the second view.
    Note that both views are constructed simultaneously.
    \item \emph{Standardize}: motivated by scaling in CV, this technique simply applies a standardization of the features by applying a \emph{z-score} standardization as ${\mX' = (\frac{\mX^T - \overline{\vx}}{\vs})^T }$,
    where $\overline{\vx} \in \mathbb{R}^{F \times 1}$ and $\vs \in \mathbb{R}^{F \times 1}$ are the mean and standard deviation vectors associated to each feature.
    Though the values are scaled, the signal encoded in $\mX'$ is equivalent to that of $\mX$, hence providing a scaled view of the original features.
    \item \emph{Local Degree Profile (LDP)}: several real-world graphs come with no features associated to nodes, and~\cite{cai2019simple} propose a mechanism for building node features based on five statistics computed from its local degree profile, $\mX' \in \sR^{N \times 5}$.
    We use this features to generate an augmented view of the graph where LDP is the node feature.
    We apply zero padding on $\mX'$, so that the feature dimension of $\mX$ and $\mX'$ match, $\mX' \in \sR^{N \times F}$ 
    \item \emph{Paste}: is a feature augmentation technique that simply combines $\mX$ with the LDP features, such that the augmented feature $\mX' \in \sR^{N \times (F + 5)}$. 
    In this case a zero padding is applied on the original features, such that $\mX \in \sR^{N \times (F + 5)}$.
\end{enumerate}

In Fig.~\ref{fig:architecture}, $aug(G)$ allows us to sample and apply augmentation on the topology or features of the graph, $G$.

\subsection{Architecture}

\begin{figure}[t!]
    \centering
    \includegraphics[scale=0.32]{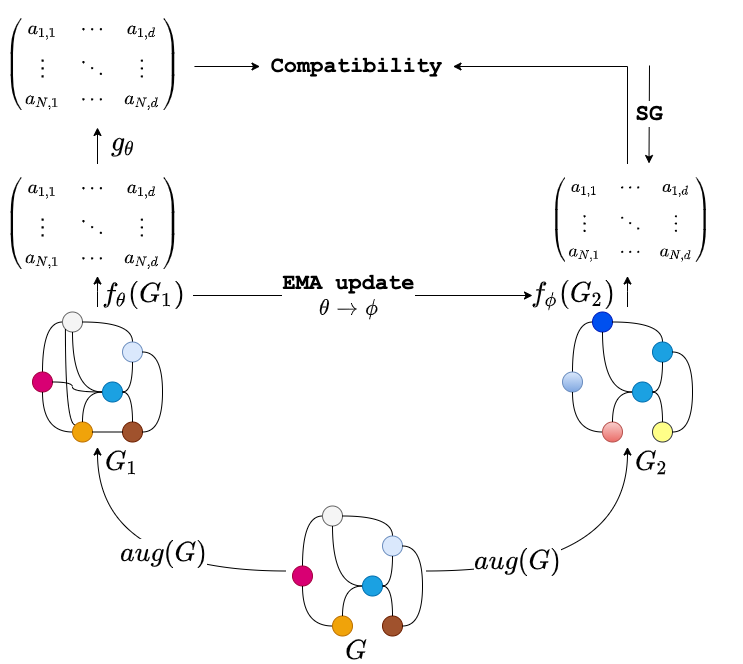}
    \caption{\model's architecture, $aug(\cdot)$ is a graph data augmentation function, $f_\theta$ and $f_\phi$ are GNN encoders parameterized by $\theta$ and $\phi$, and $g_\psi$ is a MLP, \texttt{SG} (stop gradient), and \texttt{EMA} (exponential moving average). $G_1$ and $G_2$ are the topological and feature augmentations of the graph, $G$, respectively.} 
    \label{fig:architecture}
\end{figure}

\begin{figure}[t!]
    \centering
    \includegraphics[scale=0.35]{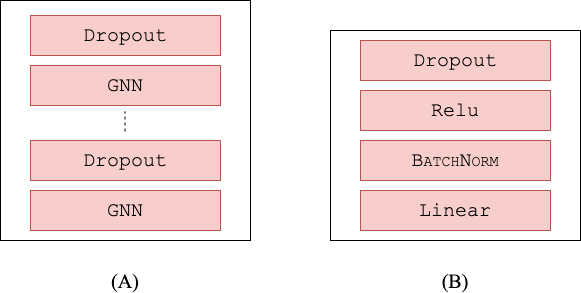}
    \caption{The architecture of (A) the encoder, $f_\cdot$, and (B) prediction or projection block, $g_\theta$. The forward propagation is from the lower to upper layers.}
    \label{fig:prediction_block}
\end{figure}

The proposed architecture, shown in Fig.~\ref{fig:architecture}, mimics a Siamese network~\cite{10.5555/2987189.2987282} that is commonly used in recent contrastive self-supervised models for representation learning~\cite{pmlr-v119-chen20j,grill2020bootstrap,chen2020exploring,robinson2021contrastive,richemond2020byol,srinivas2020curl,hassani2020contrastive}.
It has two parallel networks, referred to as a student (left hand side) and teacher (right hand side) networks~\cite{grill2020bootstrap,chen2020exploring}.
The first component of both networks is similar, they both use stacked GNN encoder functions $f_\theta$ and $f_\phi$, parameterized by the sets of parameters $\theta$ and $\phi$ (shown in Fig.~\ref{fig:prediction_block}(A)).
The encoder produces the representation of the input graph, and this is what we use for downstream graph analytic tasks. 
$f_\cdot$ is referred to as the representation block, and $\mX_1 = f_\theta(G_1), \mX_2 = f_\phi(G_2)$.
Note that the architecture is GNN encoder agnostic, thus any type of GNN can be used.

The second block in most Siamese networks~\cite{pmlr-v119-chen20j,grill2020bootstrap,chen2020exploring} is a projection head (Shown in Fig.~\ref{fig:prediction_block}(B)).
In this study, we have investigated the necessity of this head and empirically found out that for GRL it adds no qualitative gain, unlike as it is suggested for CV tasks~\cite{pmlr-v119-chen20j}.
Therefore we drop the projection head.

Finally, one key difference between the student and the teacher network is that, the student network applies a prediction block, a MLP  or the function $g_\theta$ over the output of the representation block, $f_\theta$.
Its architecture is shown in Fig.~\ref{fig:prediction_block}(B).
The prediction block act as the brain of the student who is trying to learn what is produced by the teacher network, $g_\theta(\mX_1) \approx \mX_2$

Besides their architecture, the second key difference between the two networks is that, the parameters of the student network are updated through a back-propagation of gradients.
Whereas, the parameters of the teacher network are updated using an exponential moving average (EMA) of the parameters of the student network.

Therefore, the model is trained using two joint operations, which are (1) student parameter updates using gradients and (2) teacher parameter updates using an EMA of the student parameters.
In the first case, the loss function specified using the mean squared error as in Eq.~\ref{eq:online_loss} is used to obtain the gradients.
\begin{equation}\label{eq:online_loss}
    \gL_\theta = 2 - 2 \cdot \frac{\langle g_\theta(\mX_1), \mX_2 \rangle}{\vert \vert g_\theta(\mX_1) \vert \vert_F \cdot \vert \vert \mX_2 \vert \vert_F}
\end{equation}
 
The key idea pointed out in~\cite{grill2020bootstrap} is that gradients are computed with respect to $\theta$ only, and as indicated by the $SG$ symbol on Fig.~\ref{fig:architecture}, the gradient flow on the teacher network is blocked.
Instead, the EMA of the student parameters is used as in Eq.~\ref{eq:target_update} in order to update the teacher parameters.
\begin{equation}\label{eq:target_update}
    \phi \leftarrow \tau \phi + (1 - \tau) \theta
\end{equation}
where $\tau$ is a decay rate. 

Though it seems counter-intuitive that one can learn using positive examples only, however, it has been analytically proven that the prediction block, $g_\theta$, and the {\bn} introduce implicit negative samples~\cite{tian2020understanding}.
Thus, similar to those that explicitly sample contrastive terms either from the batch~\cite{pmlr-v119-chen20j} or entire dataset~\cite{Velickovic2019DeepGI,hassani2020contrastive}, the implicit contrastive terms prevents the model from converging to the trivial solution of collapsing into a constant point.

\subsection{Improving \model}
{\model} in its current form has two efficiency problems (1) the TAs involve matrix inversion, which incurs $O(N^3)$ run time, and (2) the TAs lead to a significant increase in the number of edges.
The second problem is not favorable for full-batch GNNs in terms of GPU memory usage.
However, we propose a strategy to mitigate the severity of the aforementioned problems.

For this reason, we use subgraph sampling used in \clustergcn~\cite{10.1145/3292500.3330925}.
That is, first we create initial clusters using \metis~\cite{10.5555/305219.305248}, and then we randomly merge some clusters to create a final set of subgraphs (batches).
Then, instead of the complete graph, we apply the TAs over each subgraph independently and {\model} is trained using batches. 
This variant of {\model} is dubbed \cluster.
Section~\ref{sec:experiments} empirically corroborates the suggested improvement.


\section{Empirical Evaluation}
\label{sec:experiments}

We evaluate the performance of {\model} using publicly available real-world datasets and compare its performance against different GNN architectures and SSL baselines.

\subsection{Datasets}

\begin{table}[t!]
\begin{tabular}{|l|l|l|l|l|}
\hline
\textbf{Datasets} & \textbf{Nodes} & \textbf{Edges} & \textbf{Features} & \textbf{Classes} \\ \hline
Cora              & 2,708          & 5,278          & 1,433             & 7                \\ \hline
Citeseer          & 3,327          & 4,552          & 3,703             & 6                \\ \hline
Pubmed            & 19,717         & 44,324         & 500               & 3                \\ \hline
Photo             & 7,487          & 119,043        & 745               & 8                \\ \hline
Computers             & 13,381          & 245,778        & 767               & 10                \\ \hline
CS             & 18,333         & 81,894        & 6,805               & 15                \\ \hline
Physics             & 34,493          & 247,962        & 8,415               & 5                \\ \hline
\end{tabular}
\caption{Summary of the datasets}
\label{tbl:dataset_summary}
\end{table}

We use seven datasets provided by the PyTorch Geometric library~\footnote{\href{https://pytorch-geometric.readthedocs.io/en/latest/}{https://pytorch-geometric.readthedocs.io/en/latest/}}.
Three popular citation networks (Cora, Citeseer, Pubmed), two author collaboration networks~\cite{shchur2019pitfalls} (CS, Physics) from two categories of the Arxiv database, and co-purchased products network~\cite{shchur2019pitfalls} (Photo, Computers) from two categories of Amazon.
The summary of the datasets is given in Table~\ref{tbl:dataset_summary}.

\subsection{Baselines}

We compare the proposed method against the original semi-supervised and self-supervised GNN models.
We select three popular semi-supervised GNNs, which are \gcn~\cite{kipf2017semisupervised}, \gat~\cite{velickovic2018graph}, and \sage~\cite{10.5555/3294771.3294869}, and two self-supervised GNNs, \mvgrl~\cite{hassani2020contrastive} and \dgi~\cite{Velickovic2019DeepGI}.
Since {\mvgrl} requires topological data augmentation like ours, we use three variants {\mvgrlppr}, {\mvgrlhk}, and {\mvgrlkatz} denoting the associated augmentation techniques, which are personalized PageRank, heat kernel, and Katz-index, respectively.
A brief discussion of the baselines can be found in Section~\ref{sec:related_work}.

\subsection{Experimental Setting}

\emph{Dataset}: a public split (Planetoid split~\cite{yang2016revisiting}) of the citation networks is already available, and hence we use this split for the citation datasets.
Since no public split is available for the rest, we randomly split them as (70/10/20--train/validation/test) by respecting the class proportion in each split.

\emph{Hyperparameters}: all the algorithms are trained for 1,000 epochs, and for all of them the validation set is used to choose the best epoch for evaluating the model using the test set. 
{\model} has the same architecture and setting in all the experiments.
That is, the encoder has two GNN layers with 512 and 128 units each followed by a dropout layer, and uses the same dropout and learning rate of 0.2 and 0.0001, respectively.
Since the representation block has 128 output units, we use 128 as the representation size for all the baselines.
The hyperparameters of the topological augmentations are set to $\alpha=0.15$, $t=3$, and $\beta=0.1$.

\emph{Machine}: we use 2 NVidia Quadro RTX 5000 with NVLink, each with 3072 CUDA cores, 16 GB GDDR6 memory, and Ubuntu 19.10.

\emph{Source codes}: except {\dgi} and {\mvgrl} baselines, where we use the source code provided by the authors, for the rest of the baselines we use the PyTorch Geometric~\footnote{\href{https://pytorch-geometric.readthedocs.io/en/latest/}{https://pytorch-geometric.readthedocs.io/en/latest/}} library.

\emph{Workload}: For all the datasets the standard work-load is node classification, and the evaluation metric is accuracy.

\subsection{Results}

\begin{table*}[ht!]
\begin{tabular}{|l|ll|l|l|l|l|l|l|l|}
\hline
\multirow{3}{*}{\textbf{Dataset}} & \multicolumn{1}{l|}{\multirow{3}{*}{\textbf{GNN Type}}} & \multicolumn{8}{c|}{\textbf{Algorithms}} \\ \cline{3-10} 
 & \multicolumn{1}{l|}{} & \multirow{2}{*}{Original} & \multicolumn{7}{c|}{{\model} variants} \\ \cline{4-10} 
 & \multicolumn{1}{l|}{} &  & \ppr & \hk & \katz & \splt & \std & \paste & \ldp \\ \hline \hline
\multirow{3}{*}{\textbf{Cora}} & \multicolumn{1}{l|}{\gcn} & 79.9$\pm$0.003 & \textbf{85.3$\pm$0.02} & 85.2$\pm$0.02 & 79.9$\pm$0.02 & 84.4$\pm$0.02 & 84.6$\pm$0.03 & 81.9$\pm$0.02 & 78.9$\pm$0.02 \\ \cline{2-10} 
 & \multicolumn{1}{l|}{\gat} & 79.1$\pm$0.004 & 84.2$\pm$0.02 & 84.9$\pm$0.02 & 83.8$\pm$0.01 & 85.0$\pm$0.01 & \textbf{85.1$\pm$0.02} & 82.2$\pm$0.03 & 80.7$\pm$0.01 \\ \cline{2-10} 
 & \multicolumn{1}{l|}{\sage} & 78.7$\pm$0.01 & 84.7$\pm$0.01 & \textbf{84.9$\pm$0.01} & 79.9$\pm$0.01 & 81.7$\pm$0.01 & 84.5$\pm$0.01 & 81.9$\pm$0.01 & 68.6$\pm$0.01 \\ \hline \hline
\multirow{3}{*}{\textbf{Citeseer}} & \multicolumn{1}{l|}{\gcn} & 65.7$\pm$0.01 & 70.6$\pm$0.01 & 70.5$\pm$0.02 & 65.6$\pm$0.02 & \textbf{72.3$\pm$0.01} & 69.8$\pm$0.01 & 71.2$\pm$0.02 & 61.2$\pm$0.01 \\ \cline{2-10} 
 & \multicolumn{1}{l|}{\gat} & 66.3$\pm$0.01 & \textbf{71.4$\pm$0.02} & 71.2$\pm$0.02 & 66.5$\pm$0.03 & \textbf{71.4$\pm$0.01} & 68.0$\pm$0.03 & 70.3$\pm$0.01 & 60.1$\pm$0.02 \\ \cline{2-10} 
 & \multicolumn{1}{l|}{\sage} & 63.6$\pm$001 & 70.5$\pm$0.02 & 68.8$\pm$0.01 & 63.8$\pm$0.01 & \textbf{72.3$\pm$0.01} & 65.7$\pm$0.01 & 69.7$\pm$0.01 & 54.5$\pm$0.02 \\ \hline \hline
\multirow{3}{*}{\textbf{Pubmed}} & \multicolumn{1}{l|}{\gcn} & 79.0$\pm$0.005 & 82.1$\pm$0.01 & 82.0$\pm$0.01 & 81.0$\pm$0.01 & 81.3$\pm$0.01 & \textbf{82.3$\pm$0.01} & 76.0$\pm$0.01 & 77.1$\pm$0.01 \\ \cline{2-10} 
 & \multicolumn{1}{l|}{\gat} & 77.5$\pm$0.005 &\textbf{ 82.7$\pm$0.01} & 82.6$\pm$0.01 & 80.8$\pm$0.01 & 81.1$\pm$0.02 & 80.6$\pm$0.02 & 76.1$\pm$0.01 & 76.7$\pm$0.02 \\ \cline{2-10} 
 & \multicolumn{1}{l|}{\sage} & 76.2$\pm$0.01 & 80.2$\pm$0.01 & 80.4$\pm$0.01 & 80.7$\pm$0.02 & 79.5$\pm$0.01 & \textbf{80.8$\pm$0.02} & 72.2$\pm$0.02 & 68.6$\pm$0.02 \\ \hline \hline
\multirow{3}{*}{\textbf{Photo}} & \multicolumn{1}{l|}{\gcn} & \textbf{94.3$\pm$0.002} & 92.9$\pm$0.01 & 93.4$\pm$0.01 & 93.3$\pm$0.01 & 92.1$\pm$0.01 & 92.0$\pm$0.01 & 91.6$\pm$0.01 & 90.4$\pm$0.01 \\ \cline{2-10} 
 & \multicolumn{1}{l|}{\gat} & \textbf{94.4$\pm$0.005} & 92.7$\pm$0.01 & 93.1$\pm$0.01 & 93.2$\pm$0.01 & 93.2$\pm$0.01 & 92.2$\pm$0.01 & 91.7$\pm$0.01 & 91.7$\pm$0.01 \\ \cline{2-10} 
 & \multicolumn{1}{l|}{\sage} & \textbf{95.8$\pm$0.001} & 92.4$\pm$0.004 & 92.7$\pm$0.01 & 93.8$\pm$0.01 & 92.1$\pm$0.004 & 92.0$\pm$0.01 & 89.9$\pm$0.01 & 85.1$\pm$0.01 \\ \hline \hline
\multirow{3}{*}{\textbf{Computers}} & \multicolumn{1}{l|}{\gcn} & \textbf{90.9$\pm$0.003} & 88.6$\pm$0.01 & 88.8$\pm$0.01 & 88.6$\pm$0.01 & 88.8 $\pm$0.01 & 87.2$\pm$0.01 & 87.9$\pm$0.01 & 86.6$\pm$0.01 \\ \cline{2-10} 
 & \multicolumn{1}{l|}{\gat} & \textbf{90.9$\pm$0.004} & 88.6$\pm$0.01 & 88.8$\pm$0.01 & 88.3$\pm$0.01 & 88.6$\pm$0.01 & 88.0$\pm$0.01 & 87.6$\pm$0.01 & 87.9$\pm$0.01 \\ \cline{2-10} 
 & \multicolumn{1}{l|}{\sage} & \textbf{91.2$\pm$0.003} & 87.2$\pm$0.01 & 87.0$\pm$0.01 & 87.4$\pm$0.01 & 87.2$\pm$0.01 & 86.6$\pm$0.01 & 86.1$\pm$0.01 & 82.4$\pm$0.004 \\ \hline \hline
\multirow{3}{*}{\textbf{CS}} & \multicolumn{1}{l|}{\gcn} & \textbf{93.2$\pm$0.001} & 91.5$\pm$0.004 & 91.1$\pm$0.004 & 91.4$\pm$0.004 & 92.9$\pm$0.001 & 91.7$\pm$0.01 & 91.9$\pm$0.01 & 86.3$\pm$0.01 \\ \cline{2-10} 
 & \multicolumn{1}{l|}{\gat} & \textbf{92.8$\pm$0.002} & 91.2$\pm$0.004 & 91.4$\pm$0.004 & 91.1$\pm$0.004 & 91.9$\pm$0.01 & 90.7$\pm$0.004 & 92.3$\pm$0.01 & 85.5$\pm$0.01 \\ \cline{2-10} 
 & \multicolumn{1}{l|}{\sage} & \textbf{94.0$\pm$0.001} & -- & -- & -- & 93.0$\pm$0.001 & 90.4$\pm$0.003 & 92.9$\pm$0.004 & 86.7$\pm$0.003 \\ \hline \hline
\multirow{3}{*}{\textbf{Physics}} & \multicolumn{1}{l|}{\gcn} & \textbf{96.5$\pm$0.001} & -- & -- & -- & 95.5$\pm$0.003 & 94.8$\pm$0.0 & 95.5$\pm$0.002 & 95.0$\pm$0.002 \\ \cline{2-10} 
 & \multicolumn{1}{l|}{\gat} & \textbf{96.2$\pm$0.001} & -- & -- & -- & 95.2$\pm$0.002 & 94.6$\pm$0.002 & 95.1$\pm$0.003 & 94.6$\pm$0.003 \\ \cline{2-10} 
 & \multicolumn{1}{l|}{\sage} & -- & -- & -- & -- & -- & -- & -- & -- \\ \hline
\end{tabular}
\caption{Comparison of three (\gcn, \gat, and \sage) type of the original GNNs against the different variants of {\model} on node classification experiment. The reported results are the mean classification accuracy along with the standard deviation.
For each dataset and each GNN type, the best performing algorithm highlighted in bold. We put a blank when an algorithm fails to finish as a result of running out of GPU memory .}
\label{tbl:gnn_vs_selfgnn}
\end{table*}

\begin{table*}[ht!]
\centering
\begin{tabular}{|l|l|l|l|l|l|l|l|l|}
\hline
\multirow{2}{*}{\textbf{Dataset}} & \multicolumn{8}{c|}{\textbf{Algorithms}} \\ \cline{2-9} 
                                  & \selfbfa  & \selfppr       & \selfhk       & \multicolumn{1}{c|}{\selfkatz} & \mvgrlppr     & \mvgrlhk      & \mvgrlkatz    & \dgi          \\ \hline
\textbf{Cora}                     & \textbf{81.0$\pm$0.18} & 78.98$\pm$0.25  & 79.01$\pm$0.31 & 77.71$\pm$0.34                  & 75.3$\pm$0.72 & 74.5$\pm$0.71 & 74.5$\pm$0.6  & \textbf{81.0$\pm$0.24} \\ \hline
\textbf{Citeseer}                 & 67.19$\pm$0.37 & \textbf{68.08}$\pm$0.48  & 68.01$\pm$0.43 & 66.14$\pm$0.32                  & 58.2$\pm$0.0  & 57.5$\pm$0.07 & 61.5$\pm$0.0  & 66.3$\pm$0.28 \\ \hline
\textbf{Pubmed}                   & \textbf{80.59$\pm$0.19} & 78.05$\pm$0.15  & 77.98$\pm$0.14 & 77.40$\pm$0.17                & 73.2$\pm$0.37 & 74.9$\pm$0.43 & 74.5$\pm$0.35 & 79.8$\pm$0.19 \\ \hline
\textbf{Photo}                    & 90.86$\pm$0.1 & 93.35$\pm$0.19 & 93.38$\pm$0.16  & \textbf{93.65$\pm$0.18 }                 & 91.0$\pm$0.98 & 91.4$\pm$0.80 & 90.8$\pm$2.30 & 92.6$\pm$0.15 \\ \hline
\end{tabular}
\caption{Comparison with self-supervised GNNs. The reported results are the mean classification accuracy along with the standard deviation of 50 runs. Bold indicates the best performing algorithm. {\selfbfa} is the best of feature augmentation, and {\splt} consistently performs better.}
\label{tbl:selfgnn_vs_unsupervised_gnn}
\end{table*}
In a series of experiments, we compare {\model} against the original semi-supervised GNNs and self-supervised GNNs on node classification and report the accuracy.

\textbf{Experiment 1: Comparison with the Original GNNs}\\
In this experiment, the goal is to investigate how well {\model} compares to the original GNN models (dubbed Original).
Note, the original architectures are supervised using a small fraction of the labels, and in contrast our model has absolutely no supervision during training.
Once node embeddings are learned, we use a logistic regression classifier to classify the test set.
We use 5-fold cross validation, and the results reported in Table~\ref{tbl:gnn_vs_selfgnn} are the mean on the 40\% fold.
{\model} gives a comparable result as the original methods, in fact achieving better performance for the citation networks.

Interestingly, albeit expected, the original methods are relatively hungry for labeled data.
For the citation networks, only a fraction of the training nodes were labeled, that is a rate of 0.0563, 0.0569, 0.0030 is used for Cora, Citeseer, and Pubmed, respectively.
Whereas, for the rest of the datasets all the training nodes are labeled, fully supervised, and in these cases unsurprisingly {\model} performs lower than the original models.

Furthermore, we observe that the feature augmentation techniques lead to equivalent performance as topological augmentation techniques, except LDP in most cases and {\paste} for the Pubmed datasets.
From the topological augmentations, {\ppr} and {\hk} tend to perform better than {\katz}, except for the Photo dataset.
On the other hand, Split and Standard consistently perform better than the other feature augmentation techniques.

Finally, for the relatively large graphs (CS and Physics), where the topological augmentation produces more than a million edges, {\model} runs out of GPU memory, as indicated by the dashed entries in the table. 

\begin{figure}[t!]
    \centering
    \includegraphics[scale=0.32]{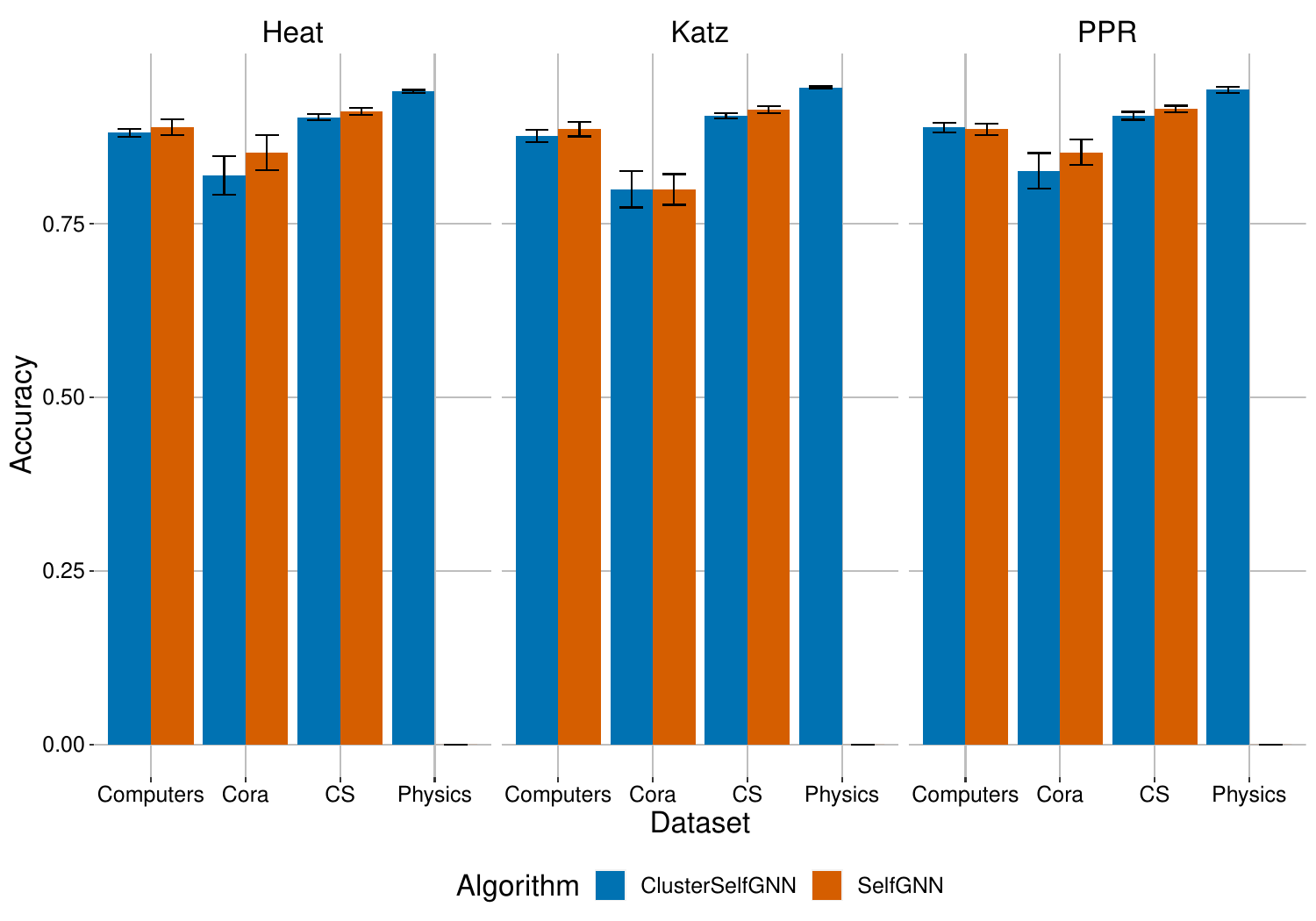}
    \caption{Performance comparison between the subgraph sampling (\cluster) and full batch (\model) based models}
    \label{fig:cluster_vs_full_batch_model}
\end{figure}

\textbf{Experiment 2: Improving Performance}\\
This experiment demonstrates the proposed remedy for the out-of memory problem pointed out in the previous experiment.
We show a comparison between the {\cluster} and {\model} to highlight the fact that {\cluster} can deal with this issue just by sacrificing a small drop in terms of performance, as shown in Fig.~\ref{fig:cluster_vs_full_batch_model}.
The comparison on the datasets other than Physics highlights that {\cluster} is closely comparable with {\model}.
However, the main purpose of {\cluster} is for relatively large graphs that do not fit in the GPU memory.
Thus, we use the Physics dataset, where {\model} runs out of memory and demonstrate \cluster's power as in the figure.
The results reported for {\cluster}, with all the three topological augmentations, are in par with the full-batch feature augmentations.

\textbf{Experiment 3: Comparison with Self-supervised GNNs}\\
Next, we perform an experiment to compare {\model} against two of the self-supervised methods {\dgi}~\cite{Velickovic2019DeepGI} and {\mvgrl}~\cite{hassani2020contrastive}.
Details on the settings of this experiment are given in Appendix~\ref{apx:settings_exp3}.
Table.~\ref{tbl:selfgnn_vs_unsupervised_gnn} reports the results of this experiment, and we can see that {\model} mostly performs better than the baselines.
In {\mvgrl} paper~\cite{hassani2020contrastive}, the reported results for Cora, Citeseer, and Pubmed are $86.8 \pm 0.5$, $73.3 \pm 0.5$, and $80.1 \pm 0.7$, respectively.
However, first the representation dimension is 512 and second it is not stated whether they use a public or their own split of the datasets.
As pointed out in a study~\cite{shchur2019pitfalls}, different splits on the three datasets lead to a remarkably huge difference in performance.

\subsection{Ablation Study}

\begin{figure}[t!]
    \centering
    \includegraphics[scale=0.33]{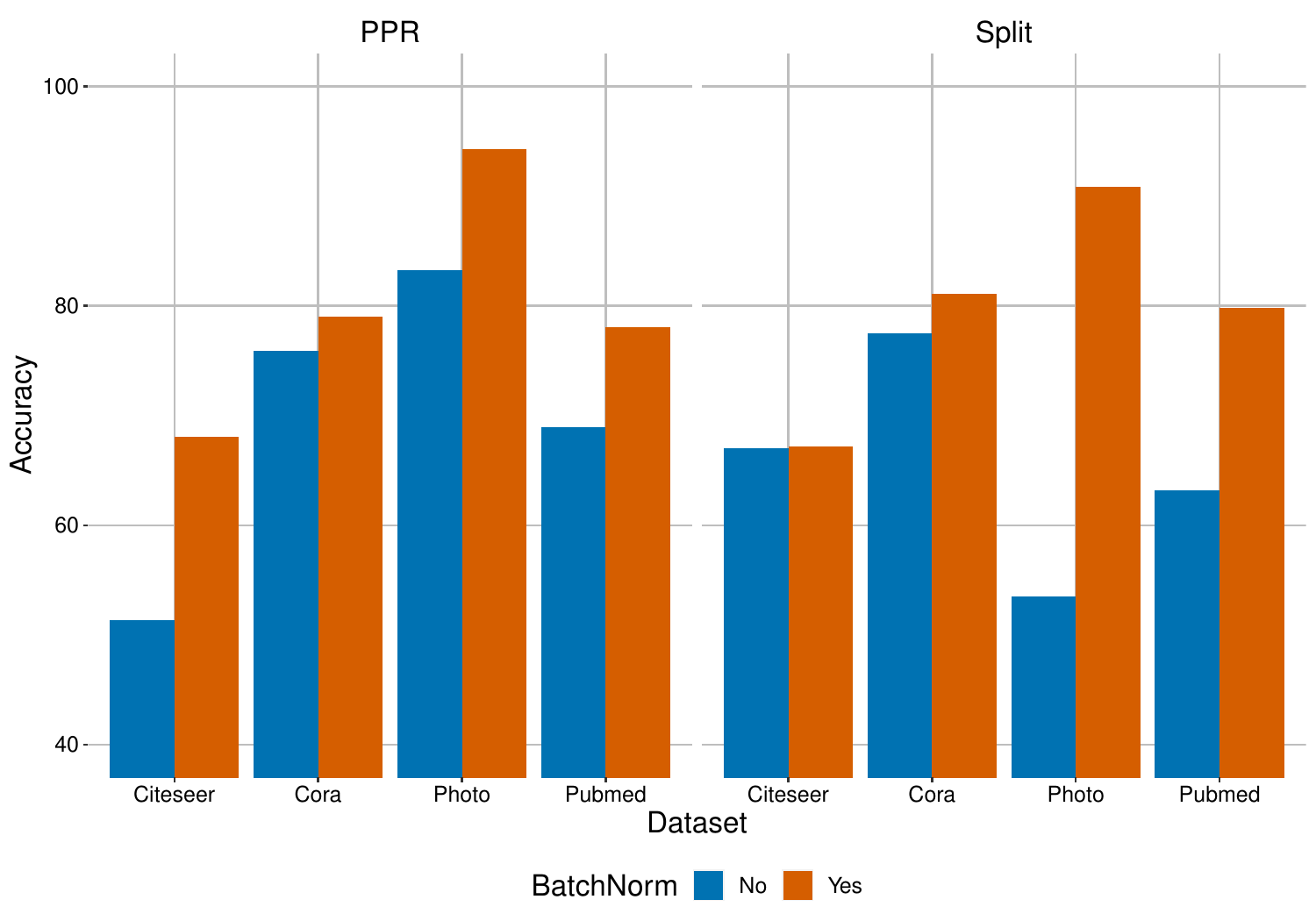}
    \caption{\model's performance with (Yes) and without (NO) {\bn}}
    \label{fig:batch_norm_effect}
\end{figure}

\begin{figure}[t!]
    \centering
    \includegraphics[scale=0.33]{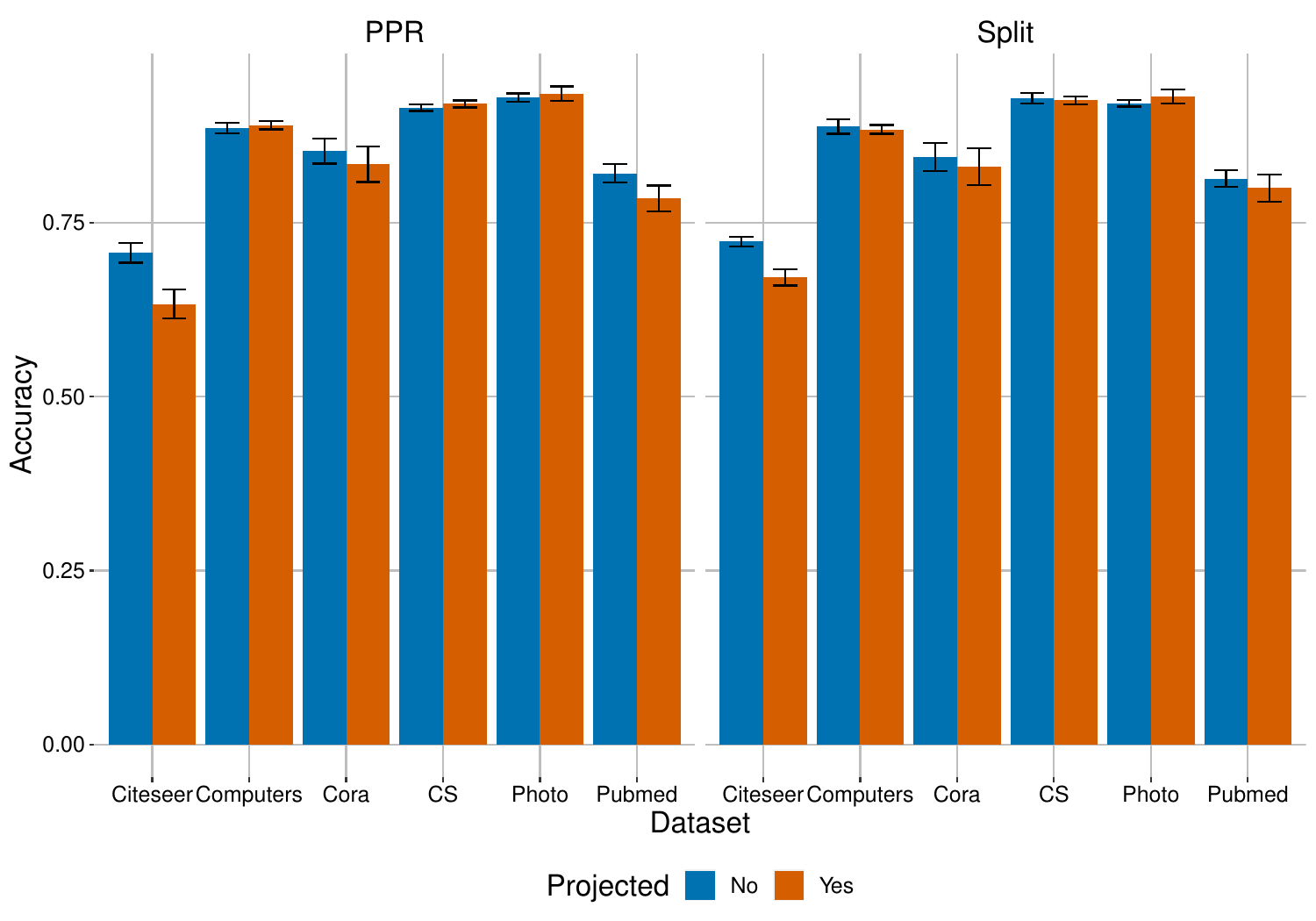}
    \caption{Comparison of {\model} with (Yes) and without (No) a projection head using the PPR and Split data augmentation techniques.}
    \label{fig:projection_head_result}
\end{figure}

\begin{figure}[ht!]
    \centering
    \includegraphics[scale=0.45]{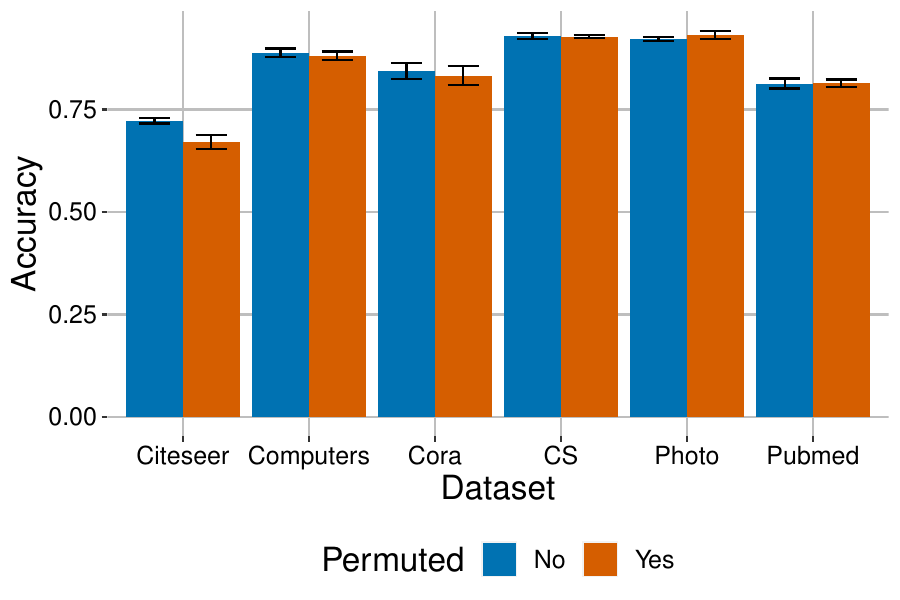}
    \caption{Analysis of the permutation of the features in the Split augmentation.}
    \label{fig:splits_results}
\end{figure}

\subsubsection{Effect of \bn}
As the key to avoiding explicit negative sampling is \bn, in the first experiment we analyze its effect on the performance of \model. 
As a result, we train {\model} with and without {\bn}.

Fig.~\ref{fig:batch_norm_effect} shows our empirical finding.
Notice that, without {\bn} \model's performance is not stable and often suffers significantly.

As an alternative to {\bn} we have investigated layer normalization (\lrn), and our finding shows that {\lrn} behaves \emph{similar} to the \emph{no} {\bn} case in Fig.~\ref{fig:batch_norm_effect}.

\subsubsection{Effect of Projection head}
A projection head is commonly used in most recent CL methods, however, \model's architecture does not make use of this head.
As a result, we use this ablation study to empirical justify our choice in removing the projection head from \model.
Fig.~\ref{fig:projection_head_result} shows the result of node classification experiment with (Yes) and without (No) a projection head.
As it can be seen from the figure, no significant improvement is achieved by adding the projection head. 
In addition, though in some cases it appears that it gives comparable performance, nonetheless it is not stable as indicated by the relatively higher variance.

\subsubsection{Effect of Perturbation on \splt}
The {\splt} augmentation requires splitting the feature along the feature dimension.
One might ask, rightly so, how sensitive this augmentation technique is to the perturbation of the features.
To this end, before splitting we apply perturbation along the feature dimension as $\mX = \mX[:, perm]$,
where $perm$ is a random permutation of the features.
Fig.~\ref{fig:splits_results} shows that overall the permutation almost has no impact on the outcome of {\model} using \splt.


\section{Conclusion and Future Work}
\label{sec:conclusion}

This study proposes \model, a novel contrastive self-supervised method for GNNs, which does not require explicit contrastive terms, negative samples.
Though negative samples are critical to the success of contrastive learning, we employ batch normalization that is shown to introduce implicit negative samples.
Furthermore, we introduce four node feature augmentation techniques that are as effective as topological ones.

We carried out extensive experiments using seven real-world datasets and show that {\model} achieves a comparable performance with supervised GNNs, while performing significantly better than semi-supervised and self-supervised methods.

{\model} relies on two parallel GNNs to be loaded into memory at the same time, this causes a major bottleneck for large networks.
Though a clustering based improvement is suggested in this study, a careful and principled work needs to be done to properly address the problem.
This is our goal for a future work.

\appendix
\section{Settings for Experiment 3}
\label{apx:settings_exp3}

\begin{table}[t!]
\begin{tabular}{|l|l|l|l|}
\hline
\textbf{Dataset} & \textbf{Augmentation} & \textbf{Dropout} & \textbf{Epochs} \\ \hline
Cora             & \splt                  & 0.35             & 1500            \\ \hline
Cora             & \ppr                   & 0.8              & 200             \\ \hline
Cora             & \hk                    & 0.7              & 200             \\ \hline
Cora             & \katz                  & 0.7              & 200             \\ \hline
Citeseer         & \splt                  & 0.2              & 1500            \\ \hline
Citeseer         & \ppr                   & 0.8              & 400             \\ \hline
Citeseer         & \hk                    & 0.7              & 400             \\ \hline
Citeseer         & \katz                  & 0.7              & 400             \\ \hline
Pubmed           & \splt                  & 0.2              & 1500            \\ \hline
Pubmed           & \ppr                   & 0.4              & 400             \\ \hline
Pubmed           & \hk                    & 0.4              & 400             \\ \hline
Pubmed           & \katz                  & 0.4              & 400             \\ \hline
Photo            & \splt                  & 0.2              & 1500            \\ \hline
Photo            & \ppr                   & 0.65             & 400             \\ \hline
Photo            & \hk                    & 0.65             & 400             \\ \hline
Photo            & \katz                  & 0.65             & 400             \\ \hline
\end{tabular}
\caption{The dropout rate and number of epochs used for experiment three}
\label{tbl:dropout_and_epochs_exp_3}
\end{table}

Experiment 3 compares {\model} against SOTA Self-supervised methods; and all of them are trained using the entire dataset.

For evaluation, we closely follow~\cite{Velickovic2019DeepGI}; that is, we train a logistic regression classifier using the training partition (The Planetoid~\cite{yang2016revisiting} partition for the three citation networks, and our own partition for Photo dataset).
The validation set is used for tuning the hyperparameters of our model.
Finally, the results reported in Table~\ref{tbl:selfgnn_vs_unsupervised_gnn} are using the test partition.
The source code provides all these details.

In Table~\ref{tbl:dropout_and_epochs_exp_3}, we provide the dropout rates and number of epochs used for this experiment.
Other than these two, all the hyperparamters are fixed.
We use a two layer GCN~\cite{kipf2017semisupervised} with 512, 128 units, and learning rate 0.0001.

We observe that topological augmentations (TA) require strong regularization and small number of epochs. On the other-hand, feature augmentation (FA) methods are gently penalized and require large number of epochs.
We conjecture that this is due to the high-order network used in TA that enables faster feature propagation than FA, which relies on the first-order network.

\bibliographystyle{ACM-Reference-Format}
\bibliography{main}


\begin{thebibliography}{35}


\ifx \showCODEN    \undefined \def \showCODEN     #1{\unskip}     \fi
\ifx \showDOI      \undefined \def \showDOI       #1{#1}\fi
\ifx \showISBNx    \undefined \def \showISBNx     #1{\unskip}     \fi
\ifx \showISBNxiii \undefined \def \showISBNxiii  #1{\unskip}     \fi
\ifx \showISSN     \undefined \def \showISSN      #1{\unskip}     \fi
\ifx \showLCCN     \undefined \def \showLCCN      #1{\unskip}     \fi
\ifx \shownote     \undefined \def \shownote      #1{#1}          \fi
\ifx \showarticletitle \undefined \def \showarticletitle #1{#1}   \fi
\ifx \showURL      \undefined \def \showURL       {\relax}        \fi
\providecommand\bibfield[2]{#2}
\providecommand\bibinfo[2]{#2}
\providecommand\natexlab[1]{#1}
\providecommand\showeprint[2][]{arXiv:#2}

\bibitem[\protect\citeauthoryear{Bromley, Guyon, LeCun, S\"{a}ckinger, and
  Shah}{Bromley et~al\mbox{.}}{1993}]%
        {10.5555/2987189.2987282}
\bibfield{author}{\bibinfo{person}{Jane Bromley}, \bibinfo{person}{Isabelle
  Guyon}, \bibinfo{person}{Yann LeCun}, \bibinfo{person}{Eduard S\"{a}ckinger},
  {and} \bibinfo{person}{Roopak Shah}.} \bibinfo{year}{1993}\natexlab{}.
\newblock \showarticletitle{Signature Verification Using a "Siamese" Time Delay
  Neural Network}. In \bibinfo{booktitle}{\emph{Proceedings of the 6th
  International Conference on Neural Information Processing Systems}} (Denver,
  Colorado) \emph{(\bibinfo{series}{NIPS'93})}. \bibinfo{publisher}{Morgan
  Kaufmann Publishers Inc.}, \bibinfo{address}{San Francisco, CA, USA},
  \bibinfo{pages}{737–744}.
\newblock


\bibitem[\protect\citeauthoryear{Brown, Mann, Ryder, Subbiah, Kaplan, Dhariwal,
  Neelakantan, Shyam, Sastry, Askell, Agarwal, Herbert-Voss, Krueger, Henighan,
  Child, Ramesh, Ziegler, Wu, Winter, Hesse, Chen, Sigler, Litwin, Gray, Chess,
  Clark, Berner, McCandlish, Radford, Sutskever, and Amodei}{Brown
  et~al\mbox{.}}{2020}]%
        {brown2020language}
\bibfield{author}{\bibinfo{person}{Tom~B. Brown}, \bibinfo{person}{Benjamin
  Mann}, \bibinfo{person}{Nick Ryder}, \bibinfo{person}{Melanie Subbiah},
  \bibinfo{person}{Jared Kaplan}, \bibinfo{person}{Prafulla Dhariwal},
  \bibinfo{person}{Arvind Neelakantan}, \bibinfo{person}{Pranav Shyam},
  \bibinfo{person}{Girish Sastry}, \bibinfo{person}{Amanda Askell},
  \bibinfo{person}{Sandhini Agarwal}, \bibinfo{person}{Ariel Herbert-Voss},
  \bibinfo{person}{Gretchen Krueger}, \bibinfo{person}{Tom Henighan},
  \bibinfo{person}{Rewon Child}, \bibinfo{person}{Aditya Ramesh},
  \bibinfo{person}{Daniel~M. Ziegler}, \bibinfo{person}{Jeffrey Wu},
  \bibinfo{person}{Clemens Winter}, \bibinfo{person}{Christopher Hesse},
  \bibinfo{person}{Mark Chen}, \bibinfo{person}{Eric Sigler},
  \bibinfo{person}{Mateusz Litwin}, \bibinfo{person}{Scott Gray},
  \bibinfo{person}{Benjamin Chess}, \bibinfo{person}{Jack Clark},
  \bibinfo{person}{Christopher Berner}, \bibinfo{person}{Sam McCandlish},
  \bibinfo{person}{Alec Radford}, \bibinfo{person}{Ilya Sutskever}, {and}
  \bibinfo{person}{Dario Amodei}.} \bibinfo{year}{2020}\natexlab{}.
\newblock \bibinfo{title}{Language Models are Few-Shot Learners}.
\newblock
\newblock
\showeprint[arxiv]{2005.14165}~[cs.CL]


\bibitem[\protect\citeauthoryear{Cai and Wang}{Cai and Wang}{2019}]%
        {cai2019simple}
\bibfield{author}{\bibinfo{person}{Chen Cai} {and} \bibinfo{person}{Yusu
  Wang}.} \bibinfo{year}{2019}\natexlab{}.
\newblock \bibinfo{title}{A simple yet effective baseline for non-attributed
  graph classification}.
\newblock
\newblock
\showeprint[arxiv]{1811.03508}~[cs.LG]


\bibitem[\protect\citeauthoryear{Chen, Ma, and Xiao}{Chen
  et~al\mbox{.}}{2018}]%
        {chen2018fastgcn}
\bibfield{author}{\bibinfo{person}{Jie Chen}, \bibinfo{person}{Tengfei Ma},
  {and} \bibinfo{person}{Cao Xiao}.} \bibinfo{year}{2018}\natexlab{}.
\newblock \bibinfo{title}{FastGCN: Fast Learning with Graph Convolutional
  Networks via Importance Sampling}.
\newblock
\newblock
\showeprint[arxiv]{1801.10247}~[cs.LG]


\bibitem[\protect\citeauthoryear{Chen, Kornblith, Norouzi, and Hinton}{Chen
  et~al\mbox{.}}{2020}]%
        {pmlr-v119-chen20j}
\bibfield{author}{\bibinfo{person}{Ting Chen}, \bibinfo{person}{Simon
  Kornblith}, \bibinfo{person}{Mohammad Norouzi}, {and}
  \bibinfo{person}{Geoffrey Hinton}.} \bibinfo{year}{2020}\natexlab{}.
\newblock \showarticletitle{A Simple Framework for Contrastive Learning of
  Visual Representations}. In \bibinfo{booktitle}{\emph{Proceedings of the 37th
  International Conference on Machine Learning}}
  \emph{(\bibinfo{series}{Proceedings of Machine Learning Research},
  Vol.~\bibinfo{volume}{119})}. \bibinfo{publisher}{PMLR},
  \bibinfo{address}{Virtual}, \bibinfo{pages}{1597--1607}.
\newblock


\bibitem[\protect\citeauthoryear{Chen and He}{Chen and He}{2020}]%
        {chen2020exploring}
\bibfield{author}{\bibinfo{person}{Xinlei Chen} {and} \bibinfo{person}{Kaiming
  He}.} \bibinfo{year}{2020}\natexlab{}.
\newblock \bibinfo{title}{Exploring Simple Siamese Representation Learning}.
\newblock
\newblock
\showeprint[arxiv]{2011.10566}~[cs.CV]


\bibitem[\protect\citeauthoryear{Chiang, Liu, Si, Li, Bengio, and Hsieh}{Chiang
  et~al\mbox{.}}{2019}]%
        {10.1145/3292500.3330925}
\bibfield{author}{\bibinfo{person}{Wei-Lin Chiang}, \bibinfo{person}{Xuanqing
  Liu}, \bibinfo{person}{Si Si}, \bibinfo{person}{Yang Li},
  \bibinfo{person}{Samy Bengio}, {and} \bibinfo{person}{Cho-Jui Hsieh}.}
  \bibinfo{year}{2019}\natexlab{}.
\newblock \bibinfo{booktitle}{\emph{Cluster-GCN: An Efficient Algorithm for
  Training Deep and Large Graph Convolutional Networks}}.
\newblock \bibinfo{publisher}{Association for Computing Machinery},
  \bibinfo{address}{New York, NY, USA}, \bibinfo{pages}{257–266}.
\newblock
\showISBNx{9781450362016}
\urldef\tempurl%
\url{https://doi.org/10.1145/3292500.3330925}
\showURL{%
\tempurl}


\bibitem[\protect\citeauthoryear{Chung}{Chung}{2007}]%
        {Chung2007TheHK}
\bibfield{author}{\bibinfo{person}{F. Chung}.} \bibinfo{year}{2007}\natexlab{}.
\newblock \showarticletitle{The heat kernel as the pagerank of a graph}.
\newblock \bibinfo{journal}{\emph{Proceedings of the National Academy of
  Sciences}}  \bibinfo{volume}{104} (\bibinfo{year}{2007}),
  \bibinfo{pages}{19735 -- 19740}.
\newblock


\bibitem[\protect\citeauthoryear{Cong, Forsati, Kandemir, and Mahdavi}{Cong
  et~al\mbox{.}}{2020}]%
        {10.1145/3394486.3403192}
\bibfield{author}{\bibinfo{person}{Weilin Cong}, \bibinfo{person}{Rana
  Forsati}, \bibinfo{person}{Mahmut Kandemir}, {and} \bibinfo{person}{Mehrdad
  Mahdavi}.} \bibinfo{year}{2020}\natexlab{}.
\newblock \bibinfo{booktitle}{\emph{Minimal Variance Sampling with Provable
  Guarantees for Fast Training of Graph Neural Networks}}.
\newblock \bibinfo{publisher}{Association for Computing Machinery},
  \bibinfo{address}{New York, NY, USA}.
\newblock
\showISBNx{9781450379984}


\bibitem[\protect\citeauthoryear{Devlin, Chang, Lee, and Toutanova}{Devlin
  et~al\mbox{.}}{2019}]%
        {devlin2019bert}
\bibfield{author}{\bibinfo{person}{Jacob Devlin}, \bibinfo{person}{Ming-Wei
  Chang}, \bibinfo{person}{Kenton Lee}, {and} \bibinfo{person}{Kristina
  Toutanova}.} \bibinfo{year}{2019}\natexlab{}.
\newblock \bibinfo{title}{BERT: Pre-training of Deep Bidirectional Transformers
  for Language Understanding}.
\newblock
\newblock
\showeprint[arxiv]{1810.04805}~[cs.CL]


\bibitem[\protect\citeauthoryear{Fetterman and Albrecht}{Fetterman and
  Albrecht}{2020}]%
        {fetterman2020understandingbyol}
\bibfield{author}{\bibinfo{person}{Abe Fetterman} {and} \bibinfo{person}{Josh
  Albrecht}.} \bibinfo{year}{2020}\natexlab{}.
\newblock \bibinfo{title}{Understanding self-supervised and contrastive
  learning with bootstrap your own latent (BYOL)}.
\newblock
\newblock
\urldef\tempurl%
\url{https://untitled-ai.github.io/understanding-self-supervisedcontrastive-learning.html}
\showURL{%
\tempurl}


\bibitem[\protect\citeauthoryear{Grill, Strub, Altché, Tallec, Richemond,
  Buchatskaya, Doersch, Pires, Guo, Azar, Piot, Kavukcuoglu, Munos, and
  Valko}{Grill et~al\mbox{.}}{2020}]%
        {grill2020bootstrap}
\bibfield{author}{\bibinfo{person}{Jean-Bastien Grill},
  \bibinfo{person}{Florian Strub}, \bibinfo{person}{Florent Altché},
  \bibinfo{person}{Corentin Tallec}, \bibinfo{person}{Pierre~H. Richemond},
  \bibinfo{person}{Elena Buchatskaya}, \bibinfo{person}{Carl Doersch},
  \bibinfo{person}{Bernardo~Avila Pires}, \bibinfo{person}{Zhaohan~Daniel Guo},
  \bibinfo{person}{Mohammad~Gheshlaghi Azar}, \bibinfo{person}{Bilal Piot},
  \bibinfo{person}{Koray Kavukcuoglu}, \bibinfo{person}{Rémi Munos}, {and}
  \bibinfo{person}{Michal Valko}.} \bibinfo{year}{2020}\natexlab{}.
\newblock \bibinfo{title}{Bootstrap your own latent: A new approach to
  self-supervised Learning}.
\newblock
\newblock
\showeprint[arxiv]{2006.07733}~[cs.LG]


\bibitem[\protect\citeauthoryear{Hamilton, Ying, and Leskovec}{Hamilton
  et~al\mbox{.}}{2017}]%
        {10.5555/3294771.3294869}
\bibfield{author}{\bibinfo{person}{William~L. Hamilton}, \bibinfo{person}{Rex
  Ying}, {and} \bibinfo{person}{Jure Leskovec}.}
  \bibinfo{year}{2017}\natexlab{}.
\newblock \showarticletitle{Inductive Representation Learning on Large Graphs}.
  In \bibinfo{booktitle}{\emph{Proceedings of the 31st International Conference
  on Neural Information Processing Systems}} (Long Beach, California, USA)
  \emph{(\bibinfo{series}{NIPS'17})}. \bibinfo{publisher}{Curran Associates
  Inc.}, \bibinfo{address}{Red Hook, NY, USA}, \bibinfo{pages}{1025–1035}.
\newblock
\showISBNx{9781510860964}


\bibitem[\protect\citeauthoryear{Hassani and Khasahmadi}{Hassani and
  Khasahmadi}{2020}]%
        {hassani2020contrastive}
\bibfield{author}{\bibinfo{person}{Kaveh Hassani} {and}
  \bibinfo{person}{Amir~Hosein Khasahmadi}.} \bibinfo{year}{2020}\natexlab{}.
\newblock \bibinfo{title}{Contrastive Multi-View Representation Learning on
  Graphs}.
\newblock
\newblock
\showeprint[arxiv]{2006.05582}~[cs.LG]


\bibitem[\protect\citeauthoryear{He, Fan, Wu, Xie, and Girshick}{He
  et~al\mbox{.}}{2020}]%
        {he2020momentum}
\bibfield{author}{\bibinfo{person}{Kaiming He}, \bibinfo{person}{Haoqi Fan},
  \bibinfo{person}{Yuxin Wu}, \bibinfo{person}{Saining Xie}, {and}
  \bibinfo{person}{Ross Girshick}.} \bibinfo{year}{2020}\natexlab{}.
\newblock \bibinfo{title}{Momentum Contrast for Unsupervised Visual
  Representation Learning}.
\newblock
\newblock
\showeprint[arxiv]{1911.05722}~[cs.CV]


\bibitem[\protect\citeauthoryear{Hu, Liu, Gomes, Zitnik, Liang, Pande, and
  Leskovec}{Hu et~al\mbox{.}}{2020b}]%
        {hu2020strategies}
\bibfield{author}{\bibinfo{person}{Weihua Hu}, \bibinfo{person}{Bowen Liu},
  \bibinfo{person}{Joseph Gomes}, \bibinfo{person}{Marinka Zitnik},
  \bibinfo{person}{Percy Liang}, \bibinfo{person}{Vijay Pande}, {and}
  \bibinfo{person}{Jure Leskovec}.} \bibinfo{year}{2020}\natexlab{b}.
\newblock \bibinfo{title}{Strategies for Pre-training Graph Neural Networks}.
\newblock
\newblock
\showeprint[arxiv]{1905.12265}~[cs.LG]


\bibitem[\protect\citeauthoryear{Hu, Dong, Wang, Chang, and Sun}{Hu
  et~al\mbox{.}}{2020a}]%
        {10.1145/3394486.3403237}
\bibfield{author}{\bibinfo{person}{Ziniu Hu}, \bibinfo{person}{Yuxiao Dong},
  \bibinfo{person}{Kuansan Wang}, \bibinfo{person}{Kai-Wei Chang}, {and}
  \bibinfo{person}{Yizhou Sun}.} \bibinfo{year}{2020}\natexlab{a}.
\newblock \bibinfo{booktitle}{\emph{GPT-GNN: Generative Pre-Training of Graph
  Neural Networks}}.
\newblock \bibinfo{publisher}{Association for Computing Machinery},
  \bibinfo{address}{New York, NY, USA}, \bibinfo{pages}{1857–1867}.
\newblock
\showISBNx{9781450379984}


\bibitem[\protect\citeauthoryear{Karypis and Kumar}{Karypis and Kumar}{1998}]%
        {10.5555/305219.305248}
\bibfield{author}{\bibinfo{person}{George Karypis} {and} \bibinfo{person}{Vipin
  Kumar}.} \bibinfo{year}{1998}\natexlab{}.
\newblock \showarticletitle{A Fast and High Quality Multilevel Scheme for
  Partitioning Irregular Graphs}.
\newblock \bibinfo{journal}{\emph{SIAM J. Sci. Comput.}} \bibinfo{volume}{20},
  \bibinfo{number}{1} (\bibinfo{date}{Dec.} \bibinfo{year}{1998}).
\newblock
\showISSN{1064-8275}


\bibitem[\protect\citeauthoryear{Kefato and Girdzijauskas}{Kefato and
  Girdzijauskas}{2020}]%
        {kefato2020gossip}
\bibfield{author}{\bibinfo{person}{Zekarias~T. Kefato} {and}
  \bibinfo{person}{Sarunas Girdzijauskas}.} \bibinfo{year}{2020}\natexlab{}.
\newblock \showarticletitle{Gossip and Attend: Context-Sensitive Graph
  Representation Learning}. In \bibinfo{booktitle}{\emph{In Proc. of the 14th
  AAAI International Conference on Web and Social Media}}
  \emph{(\bibinfo{series}{ICWSM'121})}.
\newblock
\showeprint[arxiv]{2004.00413}~[cs.LG]


\bibitem[\protect\citeauthoryear{Kipf and Welling}{Kipf and Welling}{2017}]%
        {kipf2017semisupervised}
\bibfield{author}{\bibinfo{person}{Thomas~N. Kipf} {and} \bibinfo{person}{Max
  Welling}.} \bibinfo{year}{2017}\natexlab{}.
\newblock \bibinfo{title}{Semi-Supervised Classification with Graph
  Convolutional Networks}.
\newblock
\newblock
\showeprint[arxiv]{1609.02907}~[cs.LG]


\bibitem[\protect\citeauthoryear{Klicpera, Weißenberger, and
  Günnemann}{Klicpera et~al\mbox{.}}{2019}]%
        {klicpera2019diffusion}
\bibfield{author}{\bibinfo{person}{Johannes Klicpera}, \bibinfo{person}{Stefan
  Weißenberger}, {and} \bibinfo{person}{Stephan Günnemann}.}
  \bibinfo{year}{2019}\natexlab{}.
\newblock \bibinfo{title}{Diffusion Improves Graph Learning}.
\newblock
\newblock
\showeprint[arxiv]{1911.05485}~[cs.SI]


\bibitem[\protect\citeauthoryear{Perozzi, Al-Rfou, and Skiena}{Perozzi
  et~al\mbox{.}}{2014}]%
        {10.1145/2623330.2623732}
\bibfield{author}{\bibinfo{person}{Bryan Perozzi}, \bibinfo{person}{Rami
  Al-Rfou}, {and} \bibinfo{person}{Steven Skiena}.}
  \bibinfo{year}{2014}\natexlab{}.
\newblock \showarticletitle{DeepWalk: Online Learning of Social
  Representations}. In \bibinfo{booktitle}{\emph{Proceedings of the 20th ACM
  SIGKDD International Conference on Knowledge Discovery and Data Mining}} (New
  York, New York, USA) \emph{(\bibinfo{series}{KDD '14})}.
  \bibinfo{publisher}{ACM}, \bibinfo{address}{New York, NY, USA},
  \bibinfo{pages}{701–710}.
\newblock
\showISBNx{9781450329569}


\bibitem[\protect\citeauthoryear{Qiu, Chen, Dong, Zhang, Yang, Ding, Wang, and
  Tang}{Qiu et~al\mbox{.}}{2020}]%
        {10.1145/3394486.3403168}
\bibfield{author}{\bibinfo{person}{Jiezhong Qiu}, \bibinfo{person}{Qibin Chen},
  \bibinfo{person}{Yuxiao Dong}, \bibinfo{person}{Jing Zhang},
  \bibinfo{person}{Hongxia Yang}, \bibinfo{person}{Ming Ding},
  \bibinfo{person}{Kuansan Wang}, {and} \bibinfo{person}{Jie Tang}.}
  \bibinfo{year}{2020}\natexlab{}.
\newblock \bibinfo{booktitle}{\emph{GCC: Graph Contrastive Coding for Graph
  Neural Network Pre-Training}}.
\newblock \bibinfo{publisher}{Association for Computing Machinery},
  \bibinfo{address}{New York, NY, USA}, \bibinfo{pages}{1150–1160}.
\newblock
\showISBNx{9781450379984}
\urldef\tempurl%
\url{https://doi.org/10.1145/3394486.3403168}
\showURL{%
\tempurl}


\bibitem[\protect\citeauthoryear{Richemond, Grill, Altché, Tallec, Strub,
  Brock, Smith, De, Pascanu, Piot, and Valko}{Richemond et~al\mbox{.}}{2020}]%
        {richemond2020byol}
\bibfield{author}{\bibinfo{person}{Pierre~H. Richemond},
  \bibinfo{person}{Jean-Bastien Grill}, \bibinfo{person}{Florent Altché},
  \bibinfo{person}{Corentin Tallec}, \bibinfo{person}{Florian Strub},
  \bibinfo{person}{Andrew Brock}, \bibinfo{person}{Samuel Smith},
  \bibinfo{person}{Soham De}, \bibinfo{person}{Razvan Pascanu},
  \bibinfo{person}{Bilal Piot}, {and} \bibinfo{person}{Michal Valko}.}
  \bibinfo{year}{2020}\natexlab{}.
\newblock \bibinfo{title}{BYOL works even without batch statistics}.
\newblock
\newblock
\showeprint[arxiv]{2010.10241}~[stat.ML]


\bibitem[\protect\citeauthoryear{Robinson, Chuang, Sra, and Jegelka}{Robinson
  et~al\mbox{.}}{2021}]%
        {robinson2021contrastive}
\bibfield{author}{\bibinfo{person}{Joshua Robinson}, \bibinfo{person}{Ching-Yao
  Chuang}, \bibinfo{person}{Suvrit Sra}, {and} \bibinfo{person}{Stefanie
  Jegelka}.} \bibinfo{year}{2021}\natexlab{}.
\newblock \bibinfo{title}{Contrastive Learning with Hard Negative Samples}.
\newblock
\newblock
\showeprint[arxiv]{2010.04592}~[cs.LG]


\bibitem[\protect\citeauthoryear{Shchur, Mumme, Bojchevski, and
  Günnemann}{Shchur et~al\mbox{.}}{2019}]%
        {shchur2019pitfalls}
\bibfield{author}{\bibinfo{person}{Oleksandr Shchur},
  \bibinfo{person}{Maximilian Mumme}, \bibinfo{person}{Aleksandar Bojchevski},
  {and} \bibinfo{person}{Stephan Günnemann}.} \bibinfo{year}{2019}\natexlab{}.
\newblock \bibinfo{title}{Pitfalls of Graph Neural Network Evaluation}.
\newblock
\newblock
\showeprint[arxiv]{1811.05868}~[cs.LG]


\bibitem[\protect\citeauthoryear{Srinivas, Laskin, and Abbeel}{Srinivas
  et~al\mbox{.}}{2020}]%
        {srinivas2020curl}
\bibfield{author}{\bibinfo{person}{Aravind Srinivas}, \bibinfo{person}{Michael
  Laskin}, {and} \bibinfo{person}{Pieter Abbeel}.}
  \bibinfo{year}{2020}\natexlab{}.
\newblock \bibinfo{title}{CURL: Contrastive Unsupervised Representations for
  Reinforcement Learning}.
\newblock
\newblock
\showeprint[arxiv]{2004.04136}~[cs.LG]


\bibitem[\protect\citeauthoryear{Sun, Lin, and Zhu}{Sun et~al\mbox{.}}{2020}]%
        {sun2020multistage}
\bibfield{author}{\bibinfo{person}{Ke Sun}, \bibinfo{person}{Zhouchen Lin},
  {and} \bibinfo{person}{Zhanxing Zhu}.} \bibinfo{year}{2020}\natexlab{}.
\newblock \bibinfo{title}{Multi-Stage Self-Supervised Learning for Graph
  Convolutional Networks on Graphs with Few Labels}.
\newblock
\newblock
\showeprint[arxiv]{1902.11038}~[cs.LG]


\bibitem[\protect\citeauthoryear{Tang, Qu, Wang, Zhang, Yan, and Mei}{Tang
  et~al\mbox{.}}{2015}]%
        {10.1145/2736277.2741093}
\bibfield{author}{\bibinfo{person}{Jian Tang}, \bibinfo{person}{Meng Qu},
  \bibinfo{person}{Mingzhe Wang}, \bibinfo{person}{Ming Zhang},
  \bibinfo{person}{Jun Yan}, {and} \bibinfo{person}{Qiaozhu Mei}.}
  \bibinfo{year}{2015}\natexlab{}.
\newblock \showarticletitle{LINE: Large-Scale Information Network Embedding}.
  In \bibinfo{booktitle}{\emph{Proc. of the 24th International Conference on
  World Wide Web}} (Florence, Italy) \emph{(\bibinfo{series}{WWW '15})}.
  \bibinfo{pages}{1067–1077}.
\newblock
\showISBNx{9781450334693}


\bibitem[\protect\citeauthoryear{Tian, Yu, Chen, and Ganguli}{Tian
  et~al\mbox{.}}{2020}]%
        {tian2020understanding}
\bibfield{author}{\bibinfo{person}{Yuandong Tian}, \bibinfo{person}{Lantao Yu},
  \bibinfo{person}{Xinlei Chen}, {and} \bibinfo{person}{Surya Ganguli}.}
  \bibinfo{year}{2020}\natexlab{}.
\newblock \bibinfo{title}{Understanding Self-supervised Learning with Dual Deep
  Networks}.
\newblock
\newblock
\showeprint[arxiv]{2010.00578}~[cs.LG]


\bibitem[\protect\citeauthoryear{Velickovic, Fedus, Hamilton, Li{\`o}, Bengio,
  and Hjelm}{Velickovic et~al\mbox{.}}{2019}]%
        {Velickovic2019DeepGI}
\bibfield{author}{\bibinfo{person}{Petar Velickovic}, \bibinfo{person}{W.
  Fedus}, \bibinfo{person}{William~L. Hamilton}, \bibinfo{person}{P. Li{\`o}},
  \bibinfo{person}{Yoshua Bengio}, {and} \bibinfo{person}{R.~Devon Hjelm}.}
  \bibinfo{year}{2019}\natexlab{}.
\newblock \bibinfo{title}{Deep Graph Infomax}.
\newblock
\newblock


\bibitem[\protect\citeauthoryear{Veličković, Cucurull, Casanova, Romero,
  Liò, and Bengio}{Veličković et~al\mbox{.}}{2018}]%
        {velickovic2018graph}
\bibfield{author}{\bibinfo{person}{Petar Veličković},
  \bibinfo{person}{Guillem Cucurull}, \bibinfo{person}{Arantxa Casanova},
  \bibinfo{person}{Adriana Romero}, \bibinfo{person}{Pietro Liò}, {and}
  \bibinfo{person}{Yoshua Bengio}.} \bibinfo{year}{2018}\natexlab{}.
\newblock \bibinfo{title}{Graph Attention Networks}.
\newblock
\newblock
\showeprint[arxiv]{1710.10903}~[stat.ML]


\bibitem[\protect\citeauthoryear{Yang, Cohen, and Salakhutdinov}{Yang
  et~al\mbox{.}}{2016}]%
        {yang2016revisiting}
\bibfield{author}{\bibinfo{person}{Zhilin Yang}, \bibinfo{person}{William~W.
  Cohen}, {and} \bibinfo{person}{Ruslan Salakhutdinov}.}
  \bibinfo{year}{2016}\natexlab{}.
\newblock \bibinfo{title}{Revisiting Semi-Supervised Learning with Graph
  Embeddings}.
\newblock
\newblock
\showeprint[arxiv]{1603.08861}~[cs.LG]


\bibitem[\protect\citeauthoryear{Zeng, Zhou, Srivastava, Kannan, and
  Prasanna}{Zeng et~al\mbox{.}}{2020}]%
        {zeng2020graphsaint}
\bibfield{author}{\bibinfo{person}{Hanqing Zeng}, \bibinfo{person}{Hongkuan
  Zhou}, \bibinfo{person}{Ajitesh Srivastava}, \bibinfo{person}{Rajgopal
  Kannan}, {and} \bibinfo{person}{Viktor Prasanna}.}
  \bibinfo{year}{2020}\natexlab{}.
\newblock \bibinfo{title}{GraphSAINT: Graph Sampling Based Inductive Learning
  Method}.
\newblock
\newblock
\showeprint[arxiv]{1907.04931}~[cs.LG]


\bibitem[\protect\citeauthoryear{Zou, Hu, Wang, Jiang, Sun, and Gu}{Zou
  et~al\mbox{.}}{2019}]%
        {zou2019layerdependent}
\bibfield{author}{\bibinfo{person}{Difan Zou}, \bibinfo{person}{Ziniu Hu},
  \bibinfo{person}{Yewen Wang}, \bibinfo{person}{Song Jiang},
  \bibinfo{person}{Yizhou Sun}, {and} \bibinfo{person}{Quanquan Gu}.}
  \bibinfo{year}{2019}\natexlab{}.
\newblock \bibinfo{title}{Layer-Dependent Importance Sampling for Training Deep
  and Large Graph Convolutional Networks}.
\newblock
\newblock
\showeprint[arxiv]{1911.07323}~[cs.LG]


\end{thebibliography}

\end{document}